\documentclass{article}

\usepackage[preprint]{neurips_2025}

\usepackage[utf8]{inputenc} 
\usepackage[T1]{fontenc}    
\usepackage{hyperref}       
\usepackage{url}            
\usepackage{booktabs}       
\usepackage{amsfonts}       
\usepackage{nicefrac}       
\usepackage{microtype}      
\usepackage{xcolor}         

\usepackage{amsmath, 
            amsfonts, 
            bm, 
            mathtools}

\def\eqref#1{equation~\ref{#1}}

\def\1{\bm{1}}

\DeclareMathAlphabet{\mathsfit}{\encodingdefault}{\sfdefault}{m}{sl}
\SetMathAlphabet{\mathsfit}{bold}{\encodingdefault}{\sfdefault}{bx}{n}

\def\bmu{\boldsymbol{\mu}}
\def\bSig{\boldsymbol{\Sigma}}
\def\by{\boldsymbol{y}}
\def\bz{\boldsymbol{z}}
\def\bx{\boldsymbol{x}}
\def\bth{\boldsymbol{\theta}}
\def\bph{\boldsymbol{\phi}}
\def\bps{\boldsymbol{\psi}}

\usepackage{natbib}
\usepackage{dsfont, bbm}
\usepackage{graphicx}
\usepackage{multirow}
\usepackage{wrapfig}

\title{Geometric Moment Alignment for \\ Domain Adaptation via Siegel Embeddings}

\author{%
  Shayan~Gharib\\
  Department of Computer Science\\
  University of Helsinki\\
  Helsinki, Finland\\
  \texttt{shayan.gharib@helsinki.fi}\\
  \And
  Marcelo~Hartmann\\
  Department of Computer Science\\
  University of Helsinki\\
  Helsinki, Finland\\
  \texttt{marcelo.hartmann@helsinki.fi}\\
  \AND
  Arto~Klami\\
  Department of Computer Science\\
  University of Helsinki\\
  Helsinki, Finland\\
  \texttt{arto.klami@helsinki.fi}\\
}

\newtheorem{theorem}{Theorem}
\newtheorem{definition}{Definition}
\newtheorem{remark}{Remark}
\newtheorem{proposition}{Proposition}

\begin{document}

\maketitle

\begin{abstract}
We address the problem of distribution shift in unsupervised domain adaptation with a moment-matching approach. Existing methods typically align low-order statistical moments of the source and target distributions in an embedding space using ad-hoc similarity measures. We propose a principled alternative that instead leverages the intrinsic geometry of these distributions by adopting a Riemannian distance for this alignment. Our key novelty lies in expressing the first- and second-order moments as a single symmetric positive definite (SPD) matrix through Siegel embeddings. This enables simultaneous adaptation of both moments using the natural geometric distance on the shared manifold of SPD matrices, preserving the mean and covariance structure of the source and target distributions and yielding a more faithful metric for cross-domain comparison. We connect the Riemannian manifold distance to the target-domain error bound, and validate the method on image denoising and image classification benchmarks. Our code is publicly available at \url{https://github.com/shayangharib/GeoAdapt}.
\end{abstract}

\section{Introduction}
This paper concerns a canonical machine learning (ML) challenge of improving generalization when the test condition differs from the training conditions \citep{pmlr-v97-recht19a, pmlr-v139-koh21a}. When deployed in environments that differ from the training conditions, models often suffer severe performance drops~\citep{5995347}. A key reason is distribution shift: the assumption of training and test data to follow the same distribution is rarely satisfied in practice~\citep{10.5555/1462129}. Distribution shifts can be categorized in various ways~\citep{MORENOTORRES2012521}. This paper focuses on \emph{covariate shift}, where the distribution of input features differs between the source (training) and target (test) domains, while the conditional distribution of the labels given the inputs is assumed unchanged~\citep{SHIMODAIRA2000227, JMLR:v8:sugiyama07a, NEURIPS2023_754e80f9, NEURIPS2021_4f284803}. Domain adaptation (DA) tackles this by aligning the source and target distributions, ideally without supervision. Various methods, including adversarial~\citep{pmlr-v37-ganin15, 8099799} and distance-based approaches~\citep{NIPS2016_ac627ab1}, have demonstrated success in aligning feature spaces across domains in tasks such as video~\citep{NEURIPS2021_c47e9374}, image classification~\citep{pmlr-v162-rangwani22a}, and semantic segmentation~\citep{NEURIPS2022_61aa5576}.

This paper revisits moment matching widely used for alignment of distributions in diverse applications, from style transfer~\citep{Kalischek_2021_CVPR} to inference in generative models~\citep{NEURIPS2024_3f66d5cd, zhou2025inductive}. The core idea is to align the first few moments of the source and target distributions in a shared embedding or representation space. Within DA, the early methods minimized the discrepancy in first-order statistics, most notably through maximum mean discrepancy (MMD)~\citep{pmlr-v37-long15, tzeng2014deepdomainconfusionmaximizing} with extensions exploring class-aware~\citep{ZHU2019214, 9478936, 8954037, Yan_2017_CVPR} or joint variants~\citep{pmlr-v70-long17a}. Improved alignment can be achieved by considering 
second-order statistics, by matching covariance using linear~\citep{Sun_Feng_Saenko_2016} or non-linear~\citep{10.1007/978-3-319-49409-8_35} transformations, with extensions accounting for feature discriminability~\citep{Chen_Chen_Jiang_Jin_2019}. Additionally, higher-order moments or cumulants to capture richer dependencies have been considered~\citep{ZELLINGER2019174, Chen_Fu_Chen_Jin_Cheng_Jin_Hua_2020}. 
Besides the choice of the moments, we also need to consider how the similarity is evaluated -- common to all of these methods is that they all resort to heuristic choices of the similarity, most commonly using simply the Euclidean distance between the moments. 
%
%
\begin{wrapfigure}[26]{r}{0.5\textwidth}
\centering
\includegraphics[scale = 0.65]{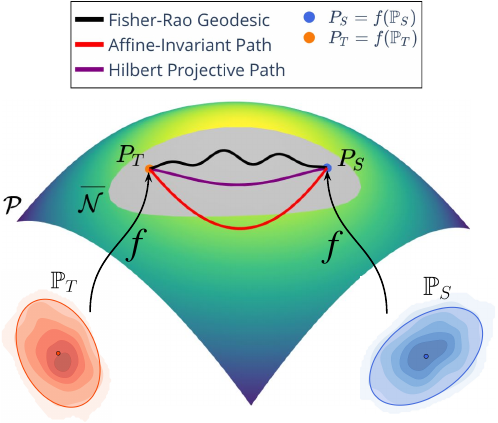}
\caption{$\mathcal{P}$ is the set of all positive-definite matrices endowed with the affine-invariant metric $g_A$. The source and target distributions $\mathbb{P}_S$ and $\mathbb{P}_T$ in the original space are pushed to $\mathcal{P}$ using the embedding $f$ and denoted as $P_S$ and $P_T$ respectively. $\overline{\mathcal{N}}$ (grey area) is a submanifold of $\mathcal{P}$ formed by the projection of Gaussians via $f$. The colored lines conceptually depict paths between them on $\mathcal{P}$: The affine-invariant path is the geodesic path (shortest) in $\mathcal{P}$, the Fisher-Rao path here is the projection by $f$ of the geodesic path on the manifold of Gaussians to $\mathcal{P}$, and the Hilbert projective path is an approximation of the affine-invariant path on $\mathcal{P}$.}
\label{fig:teaser}
\end{wrapfigure}

Riemannian geometry has been increasingly used in ML, adapting various methods for spaces more general than Euclidean; see, for example, \citet{absil:2008}, \citet{bronstein:2017}, \citet{nickel:2017}, \citet{brooks:2019} and \citet{miolane:2020}. In particular, covariances are elements of the symmetric positive-definite space (SPD), which admits a non-Euclidean geometry that better represents the eigen-structure of the problem and introduces notions of invariance~\citep{Pennec2006, doi:10.1137/050637996, Bhatia+2007}. This perspective has enabled principled algorithms for SPD-valued data, ranging from kernel methods and dimensionality-reduction on SPD manifolds to end-to-end neural architectures, and SPD manifold optimization~\citep{Jayasumana_2013_CVPR, 10.1007/978-3-319-10605-2_2, NIPS2014_3000e56b, Huang_Van_Gool_2017}. Information geometry, in particular, offers a Riemannian perspective that emphasizes the Fisher-Rao geometry on the space of probability models. This notion has allowed efficient optimization techniques, such as the natural gradient, which has been widely studied and applied in the ML context~\citep{amari:1998, martens:2020}.

Motivated by these works, some moment-matching DA methods have replaced ad-hoc Euclidean distances with geometry-aware alternatives. \citet{morerio2018minimalentropy} adopt practical approximations to SPD geometry (e.g., log-Euclidean metrics on covariances), \citet{8434290} embed covariances into a reproducing kernel Hilbert space, and \citet{Luo_Ren_Ge_Huang_Yu_2020} compare orthogonal bases of covariances via Frobenius norms. \citet{9522884} proposed mapping the features to spheres with geodesic kernels, and \citet{NEURIPS2022_28ef7ee7} integrated SPD-aware normalization and layers into the embedding network. Although these methods move beyond naive Euclidean matching and demonstrate the value of proper metrics, they either rely on surrogate spaces, discard crucial covariance information (e.g., singular values), or limit scalability by imposing specifically designed architecture for SPD matrix operations, and thus fall short in terms of practicality and efficiency. In this paper, we focus specifically on the question of how similarities should be computed and how to best transform the moments. For this, we leverage on concepts from differential geometry. We map the latent representations of both domains using a diffeomorphic transformation into the SPD manifold~\cite{CALVO1990223} (see Fig~\ref{fig:teaser}). This transformation captures the first two moments into a single SPD matrix. We then exploit the Riemannian structure of the SPD manifold to measure the distance using two geometrically inspired distances on the SPD manifold: Affine-Invariant Riemannian~\citep{Bhatia+2007} and Hilbert projective distance~\citep{pmlr-v221-nielsen23b} that approximates it. These distances can be effectively computed to quantify the discrepancy between the mapped source and target embeddings through their estimated statistical moments. We iteratively minimize this distance with respect to the parameters of a neural network using a gradient-based optimization method. In addition, we show that minimizing the Hilbert projective distance provides an upper bound on the target domain error, building on the results of~\citep{pmlr-v97-zhao19a} and~\citep{Ben-David2010}.

\section{Background}

\subsection{Problem Setup}
\label{sec:problem_setup}

Let us denote $\mathcal{X}_S, \mathcal{Y}_S$ as the input and output space of the source domain, and $\mathcal{X}_T, \mathcal{Y}_T$ as the input and output space of the target domain. Let $\mathcal{Z}$ denote the latent representation space. A feature encoder is a function $e_{\bth}: \mathcal{X} \rightarrow \mathcal{Z}$ indexed by a vector of parameters $\bth$, which transforms each input $\bx$ into latent representations $\bz$. According to the unsupervised domain adaptation (UDA) setting, we are given a labeled source domain dataset $\{ \bx_{i, S}, y_{i, S} \}_{i=1}^{n_S} \subset \mathcal{X}_S \times \mathcal{Y}_S$ and an unlabeled target domain dataset $\{ \bx_{i, T} \}_{i = 1}^{n_T} \subset \mathcal{X}_T$. We assume a covariate shift setting~\citep{SHIMODAIRA2000227}: 
\begin{align*}
p_S(\bx) \ne p_T(\bx) \quad \text{and} \quad \bar{p}_S(\by \mid \bx) = \bar{p}_T(\by \mid \bx) \ \ \forall \bx, \by.
\end{align*}
Here $p_S : \mathcal{X}_S \rightarrow \mathbb{R}^+$ and $p_T : \mathcal{X}_T \rightarrow \mathbb{R}^+$ are  probability distributions in the input spaces, and $\bar{p}_S, \bar{p}_T$ denote the conditional distributions. We assume $\mathcal{X}_S, \mathcal{X}_T \subset \mathcal{X}$ and $\mathcal{Y}_S, \mathcal{Y}_T \subset \mathcal{Y}$.

The goal is to learn simultaneously an encoder $e_{\bth}(\cdot)$ and a down-stream model, so that the performance of the model is maximized on the target domain. That is, we want $\mathcal{Z}$ that is both invariant of the domain and informative about the task of interest. The adaptation process is always  unsupervised -- we do not assume any $y_T \in \mathcal{Y}_T$ -- the task of interest can be arbitrary. We consider two examples:
\begin{itemize}
    \item \textbf{Supervised Task (ST):}
    Classification with labeled source domain, solved by simultaneous learning of the encoder $e_{\bth}$ and a label predictor $c_{\bph}: \mathcal{Z} \rightarrow \mathcal{Y}$ parameterized by $\bph$ to maximize accuracy on the target domain.

    \item \textbf{Unsupervised Task (UT):} Denoising with only the input spaces $\mathcal{X}_S$, $\mathcal{X}_T$. The encoder $e_{\bth}$ forms a compact representation in $\mathcal{Z}$ and a decoder $d_{\bps}: \mathcal{Z} \rightarrow \mathcal{X}$ parameterized by $\bps$ maps them back to the input space. The goal is to denoise target domain samples.
\end{itemize}

\subsection{Moment Matching for DA}
\label{sec:problem_setup}
Similar to prior moment matching methods, we compare empirical feature distributions to align the source and target domains in $\mathcal{Z}$. Let $\bz_{i, S} = e_{\bth}(\bx_{i, S})$ and $\bz_{i, T} = e_{\bth}(\bx_{i, T})$ denote the encoded representations of the source and target inputs, respectively. 
For the source domain the empirical first and second moments estimated from a mini-bactch of size $b_S$ are
\begin{align*}
    \bmu_S = \frac{1}{b_S}\sum_{i = 1}^{b_S} \bz_{i, S},
    \qquad
    \bSig_S = \frac{1}{b_S - 1}\sum_{i = 1}^{b_S}
    \big(\bz_{i, S} - \bmu_S \big) \big(\bz_{i, S} - \bmu_S \big)^\top,
\end{align*}
with analogous $\bmu_T$ and $\bSig_T$ for the target domain. These moment statistics serve as foundational components in our method, and following the common practice we adapt them by end-to-end training of a combined objective
\begin{equation}
\label{eq:loss_factorization}
    \mathcal{L} = \mathcal{L}_{\text{task}} + \beta \mathcal{L}_{\text{dist}},
\end{equation}
where $\mathcal{L}_{\text{task}}$ is any task-specific objective and $\mathcal{L}_{\text{dist}}$ measures the domain shift. Section~\ref{sec:method} will detail how we form $\mathcal{L}_{\text{dist}}$ that will be defined using the previous first- and second-order sample moments.

\subsection{Riemannian manifolds and information geometry} 
We review basic notions of Riemannian manifold and information geometry necessary in this work. For more details see for example \citet{docarmo:1992} and \citet{docarmo:2016}. A set $M$ is called {\it manifold} of dimension $D$ if together with bijective smooth mappings (at times called parametrization) $\varphi_i : \Theta_i \subseteq \mathbb{R}^D \rightarrow M$ satisfies (a) $\cup_i \varphi_i(\Theta_i) = M$ and (b) for each $i$, $j$ $\varphi_i(\Theta_i) \cap \varphi_j(\Theta_j) \neq \emptyset$. A manifold $M$ is called a Riemmanian manifold when it is characterized by the pair $(M, g)$ where for each $p \in M$ the metric function $g_p : T_pM \times T_pM \rightarrow \mathbb{R}$ is smooth (in $p$) and positive-definite, and associates the usual dot product of vectors in the tangent space $T_pM$ at $p$, that is $(V, U) \xrightarrow[]{g_p} g_p(V, U)$. The conditions (i) and (ii) together with the choice of $g_p$ are important because we can map a point in an open set of the Euclidean space and map it to $M$ in a diffeomorphic manner. This means that the classical tools of differential calculus on $\mathbb{R}^D$ can be used to generalize notions of differentiation to domains more general than Euclidean, and the function $g_p$ gives us a way to generalize measures of distance, angles, and areas on $M$. 

As an example, the SPD space that we use is formally defined as $\mathcal{P}(D) = \left\lbrace \bSig \in \mathbb{R}^{D \times D} : \bSig = \bSig^\top, \|\bx\|^2_{\bSig} > 0, \ \forall \bx \in \mathbb{R}^D \ \textrm{and} \ \bx \neq \boldsymbol{0} \right\rbrace$ with an explicit global parametrization found in \citet{kurowicka:2003}. Once $g_p$ has been chosen, a Riemannian distance function $d : \mathcal{P}(D) \times \mathcal{P}(D) \rightarrow [0, \infty)$ ensues. For given $\boldsymbol{q}, \boldsymbol{p} \in \mathcal{P}(D)$, there is a unique path joining $\boldsymbol{q}, \boldsymbol{p}$ whose trace now lies completely on $\mathcal{P}(D)$, and so the distance measure $d$ over $\mathcal{P}(D)$ makes sense \citep[recall][for illustrations of $\mathcal{P}(D)$]{peter:1994}. The field of information geometry studies the intrinsic geometry of the family of probability models specified by a natural choice of the function $g_p$ given by the Fisher-Rao metric. This metric is related with asymptotic statistical inference through the Crámer-Rao lower bound, and because of that there has been a great interest in understanding its properties from the differential geometry viewpoint. See \citet{kass::1997}, \citet{amari2000methods} and \citet{calin:2014} for more technical details.

\section{Method}
\label{sec:method}

\paragraph{Motivation}
The purpose of $\mathcal{L}_{\text{dist}}$ in DA is to measure the true distance between the source and the target distributions in the latent space. When juxtaposing the previous notions on Riemannian geometry with the DA goal, it seems rather appealing to pick a metric $g_p$ so that the associated Riemannian distance $d$ plays the role of a loss function $\mathcal{L}_{\text{dist}}$,  respecting the underlying geometry of the probability distributions involved. The choice of $g_p$ as the Fisher-Rao is considered optimal in the information geometry literature when the distributions belong to a parametric family. Now, however, the distributions are unknown, but we assume their first-order and second-order moments (mean and covariances) to exist and hence be available as a parameterization. That is, we need a metric $g_p$ that is a function of both the first and second moments. 

An immediate choice is the Fisher-Rao metric associated with the family of multivariate Gaussian distributions \citep{skovgaard:1984}. The corresponding distance is not known in closed-form, but many approximations have been proposed; see \citet{CALVO1990223}, \citet{pinele:2020} and \citet{e25040654}. We choose the approach proposed by \citet{CALVO1990223}, based on embeddings into the Siegel-group, whose closed-form distances on SPD spaces are known and bound the Riemannian distance with the Fisher-Rao metric \citep{e25040654}. We make two important observations regarding the choice: \textbf{1)} From the information geometry viewpoint, the Fisher-Rao metric is an optimal choice for the family of parametric distributions, for example multivariate Gaussians. However, from a pure Riemannian geometry notion, the metric can be chosen freely as long as it satisfies the smooth and positive-definite conditions \citep{petersen:2016}, making this choice valid for any family distributions --- we just characterize the distributions, and therefore, distances only in terms of the moments. \textbf{2)} 
The Riemannian distance associated with the Fisher-Rao metric in multivariate Gaussian models can also be computed, but not efficiently so that it could be used within a DA algorithm. The approximations are necessary for a practical method and, in fact, do not incur notable additional computation over the Euclidean distance.

A practical method building on this motivation is characterized next. We first transform the first two moment statistics to embed them into a submanifold on the SPD space. We then introduce a native and geometrically valid distance on the SPD space to measure the distance between the embedded distributions, and provide also a faster approximation. Finally, we prove that minimizing the approximate distance minimizes also the domain generalization error.

\subsection{Siegel Embeddings}
\label{sec:embeddings}
Our method is constructed upon the adaptation of the first two moments. For this, it is convenient to have a joint representation of both that allows us simultaneously addressing them during the adaptation process. This is achieved by the Siegel embeddings as follows.
\begin{definition}
Let $\mathcal{P}(n+1)$ denote the space of SPD matrices with dimension $(n + 1)$ and $P \in \mathcal{P}(n + 1)$ an element of it. \citet{CALVO1990223} proposed a family of diffeomorphic embeddings $f_a : \mathbb{R}^n \times \mathcal{P}(n) \rightarrow \mathcal{P}(n + 1)$ with $a > 0$ given by, 
\begin{align*}
    (\bmu, \bSig) &\stackrel{f_a}{\mapsto}
    \begin{bmatrix}
        \bSig + a\bmu \bmu^\top & a\bmu \\
        a\bmu^\top             & a
    \end{bmatrix} = P.
\end{align*}
\end{definition}
The choice of a specific $a$ defines a particular embedding within this family and effectively scales the contribution of the mean vector to the overall SPD matrix representation.
\begin{remark}
\label{remark:f_1}
For the choice of $a=1$, the family of diffeomorphic embeddings $f_a$ simplifies to a canonical form
\begin{equation}
\label{eq:f1}
    f_1(\bmu, \bSig) =
    \begin{bmatrix}
        \bSig + \bmu \bmu^\top & \bmu \\
        \bmu^\top             & 1
    \end{bmatrix}.
\end{equation}
\end{remark}
This particular mapping is central to this work. As observed by \citet{CALVO1990223}, it isometrically embeds a Gaussian manifold equipped with the Fisher metric $(\mathcal{N}(n), g_F)$ into the SPD manifold equipped with the affine-invariant metric $(\mathcal{P}(n + 1), \frac{1}{2} g_A)$. Here, the $n$-dimensional Gaussian family is denoted as $\mathcal{N}(n) = \{ \mathcal{N}_n(\bmu, \bSig) : (\bmu, \bSig) \in \mathbb{R}^n \times \mathcal{P}(n) \}$ and the affine-invariant metric is the function $g_A : T_P \mathcal{P}(n + 1) \times T_P \mathcal{P}(n + 1) \rightarrow \mathbb{R}$ given by
\begin{align*}
    (\boldsymbol{V}_1, \boldsymbol{V}_2) &\stackrel{g_A}{\mapsto} \mathrm{tr}(P^{-1}\boldsymbol{V}_1P^{-1}\boldsymbol{V}_2)
\end{align*}
where $\boldsymbol{V}_1$ and $\boldsymbol{V}_2$ are real symmetric matrices. In the following, we detail the associated Riemannian distance to $g_A$ and the implications of this embedding. From now on, we denote $f_1$ as $f$.

\subsubsection{Distance}
As mentioned above, the embedding function $f$ allows us to look at the distributions in $\mathcal{N}(n)$ as points $(\bmu, \bSig)$ on the SPD manifold $\mathcal{P}(n+1)$.
The Riemannian distance associated with the Fisher-Rao metric $g_F$ lacks a general
closed-form solution \citep{skovgaard:1984}, but it has a natural counterpart on the SPD space that has closed-form expression, characterized next.
Given two points $P_1 = f(N(\bmu_1, \bSig_1))$ and $P_2 = f(N(\bmu_2, \bSig_2))$, we use the associated Riemannian distance of the manifold $(\mathcal{P}(n + 1), \frac{1}{2} g_A)$. This Riemannian distance is given in closed form, and it also respects the geometry of the set $\mathcal{P}(n+1)$~\citep{peter:1994} and lower bounds the Fisher-Rao distance. We formalize these properties in the following.
\begin{definition}[Affine-Invariant Riemannian Distance]
Let $\big( \mathcal{P}(n+1), \tfrac{1}{2}g_A \big)$ denote the SPD space endowed with the affine-invariant metric. Given $P_1, P_2 \in \mathcal{P}(n+1)$, the Riemannian distance between any two points on this manifold is given by~\citep{Pennec2006},
\begin{equation}
\label{eq:airm_dist}
    d_A(P_1, P_2) = \left\|\text{Log}(P_1^{-1/2}P_2P_1^{-1/2})\right\|_{\mathcal{F}} = \sqrt{\frac{1}{2}\sum_{i=1}^{n+1}\log^2 \lambda_i(U)}
\end{equation}
where $\|.\|_\mathcal{F}$ is the Frobenius norm, $\text{Log}(.)$ is the matrix logarithm, $\lambda_i(U)$ is the $i$-th eigenvalue of the matrix $U$, and $U=P_1^{-1}P_2$.
\end{definition}
\begin{proposition}
~\label{prop:lower_bound}
Let $(\mathcal{N}(n), g_F)$ and $(\mathcal{P}(n+1), \tfrac{1}{2}g_A)$ be manifolds as above. \citet{CALVO1990223} showed that for any two distributions $N_1 := N_1(\bmu_1, \bSig_1), N_2 := N_2(\bmu_2, \bSig_2) \in \mathcal{N}(n)$, the distance $d_A$ between their embeddings via $f$ provides a lower bound to the Riemannian distance associated with the Fisher-Rao metric $g_F$,
\begin{equation}
    d_A(f(N_1), f(N_2)) \le d_F(N_1, N_2).
\end{equation}
\end{proposition}
where $d_F$ is the Riemannian (Fisher-Rao) distance.
\begin{remark}
The particular $f:\mathcal{N}(n)\rightarrow\mathcal{P}(n+1)$ isometrically embeds $(\mathcal{N}(n), g_F)$ into $(\mathcal{P}(n + 1), \frac{1}{2} g_A)$. This means that the metric tensor $g_F$, on $\mathcal{N}(n)$, is perfectly preserved on its image in the embedded submanifold $f(\mathcal{N}(n)) := \overline{\mathcal{N}} (n) \subset \mathcal{P}(n + 1)$. The intrinsic geodesic distance within $\overline{\mathcal{N}}$ is therefore precisely the Fisher-Rao distance. However, the submanifold $\overline{\mathcal{N}}$ is not totally geodesic within the SPD space $\mathcal{P}(n+1)$. This implies that the shortest path between two points in $\overline{\mathcal{N}}$, as judged by the metric $\tfrac{1}{2} g_A$, may exit and re-enter $\overline{\mathcal{N}}$. Consequently, this path in $(\mathcal{P}(n + 1), \frac{1}{2} g_A)$ provides a shorter or equal length to the path constrained to lie entirely within $\overline{\mathcal{N}}$, which yields the inequality in Proposition~\ref{prop:lower_bound}.
\end{remark}

The distance $d_A$ requires all eigenvalues of the matrix $U$, which may cause problems in higher dimensions. This can be avoided by considering alternative natural distance on the submanifold of embedded Gaussians within the SPD manifold. \citet{pmlr-v221-nielsen23b, e25040654} proposed the Hilbert projective distance as a computationally efficient approximation to the $d_A$ distance on $(\mathcal{P}(n+1), \tfrac{1}{2}g_A)$. Unlike the affine-invariant Riemannian distance, it depends only on the largest and smallest eigenvalues of the generalized eigenvalue problem, which can be efficiently approximated using fast iterative methods~\citep{doi:10.1137/S1064827500366124, golub13}.
\begin{definition}[Hilbert Projective Distance]
    For two SPD matrices $P_1, P_2 \in \mathcal{P}(n+1)$, the Hilbert projective distance is defined as:
    \begin{equation}
    \label{eq:hilbert_dist}
        d_{H}(P_1, P_2) = \log\left(\frac{\lambda_{\text{max}}(P_1^{-1}P_2)}{\lambda_{\text{min}}(P_1^{-1}P_2)}\right)
    \end{equation}
    where $\lambda_{\text{min}}$ and $\lambda_{\text{max}}$ are the minimum and maximum eigenvalues respectively.
\end{definition}

Therefore, we have two distance candidates $d_A$ and $d_H$ for replacing $\mathcal{L}_{\text{dist}}$ in practice:
\begin{align*}
    \min_{\boldsymbol{\theta}} \mathcal{L}_{\text{dist}}
    (\boldsymbol{\theta})
    := \min_{\boldsymbol{\theta}} d_A \Big( 
    f 
    (\bmu_S, \bSig_S) 
    ,
    f 
    (\bmu_T, \bSig_T)
    \Big)
\end{align*}
with a similar formulation for $d_H$. On the right-hand side of the above minimization problem, the Riemannian distance function is a function of $\boldsymbol{\theta}$.

\subsubsection{Theoretical Guarantee}
In this section, we provide a theoretical justification for the use of the above distances within DA. For the Hilbert projective distance (HPD) in Eq.~\ref{eq:hilbert_dist}, we will provide an upper bound for the generalization error in Theorem~\ref{thr:target_error}, whereas for the Affine-Invariant Riemannian Distance (AIRD) in Eq.~\ref{eq:airm_dist}, we established that it is bounded by the true Fisher-Rao distance. Even though we establish a formal bound only for HPD, it approximates AIRD well \citep{pmlr-v221-nielsen23b} and the direct minimization of this true metric, rather than its approximation, is intuitively very reasonable.

We start by noting that an upper bound for the target domain error is well established in the DA literature \citep{Ben-David2010, pmlr-v97-zhao19a}, combining the source error and the domain change. We show that minimizing the HPD between the source and target distributions minimizes this established upper bound, extending the results of \citet{Ben-David2010, pmlr-v97-zhao19a}. We relate the HPD to the $\tilde{\mathcal{H}}$-divergence, for which an upper bound already exists through the total variation ($TV$) divergence~\citep{Ben-David2010}. Moreover, \citet{cohen2024hyperboliccontractivityhilbertmetric} show that the $TV$-divergence is itself bounded by the HPD. Combining these results leads to our main theorem. A complete proof is provided in Appendix~\ref{proof:theorem}. 
\begin{theorem}[Upper Bound on Target Error] 
\label{thr:target_error}
Let $\mathbb{P}_S$ and $\mathbb{P}_T$ be the probability measures of the inputs in the input space for the source and target domains, and $p_S$, $p_T$ their respective density functions. Let $\gamma$ be a measure of distance between the labeling functions of the domains. For any hypothesis $h \in \mathcal{H}$, the expected error on the target domain, $\varepsilon_T(h)$, is bounded by
\begin{equation}
    \varepsilon_T(h) \leq \varepsilon_S(h) + 2\tanh\frac{d_H(\mathbb{P}_S, \mathbb{P}_T)}{4} + \gamma
\end{equation}
\end{theorem}
In this work we consider domain shift scenarios where $\gamma=0$, but note that when it is not negligible the adaptation should address also that part of the shift \citep{pmlr-v97-zhao19a}; minimizing $d_H$ or $d_A$ alone will not be sufficient. This holds for any method, not just ours.

\subsection{Computational stability}
\label{sec:practicality}
Our distances Eq.~\ref{eq:airm_dist} and Eq.~\ref{eq:hilbert_dist} involve matrix inverses, which requires ensuring invertibility of the underlying matrices throughout training. From a computational perspective, this is not an issue as computing the inverse or the eigenvalues is not a dominant factor; in all our our experiments the computational cost of both the proposed methods and all baselines are within approximately $20\%$ of each other. However, we need to ensure that $P_S$ is always invertible. The Schur complement~\citep{bernstein2009matrix} for block matrices, as in Proposition~\ref{prop:schur}, allows re-casting this requirement in terms of the covariance $\bSig$ instead. From Eq.~\ref{eq:f1} we have $\boldsymbol{A}-\boldsymbol{B}\boldsymbol{D}^{-1}\boldsymbol{C} = \bSig + \bmu\bmu^\top - \bmu\bmu^\top = \bSig$.
\begin{proposition}
\label{prop:schur}
    Let $\boldsymbol{A} \in \mathbb{R}^{n\times n}$, $\boldsymbol{B} \in \mathbb{R}^{n\times n'}$, $\boldsymbol{C} \in \mathbb{R}^{n'\times n}$, $\boldsymbol{D} \in \mathbb{R}^{n'\times n'}$. The matrix $\boldsymbol{M} = \begin{bmatrix} \boldsymbol{A} & \boldsymbol{B} \\ \boldsymbol{C} & \boldsymbol{D} \end{bmatrix}$ is then invertible if and only if $\boldsymbol{D}$ and $\boldsymbol{A} - \boldsymbol{B} \boldsymbol{D}^{-1}\boldsymbol{C}$ are non-singular.
\end{proposition}

In our experiments, we ensure this using a combination of two elements. First, we restrict the choice of the embedding space dimensionality $n$ relative to the mini-batch size $b_S$, so that $b_S \gg n$. Second, we learn the model in two phases: First we optimize only the task objective using the source data while monitoring the determinant of $P_S$, only turning the adaptation on ($\beta > 0$) once it is above a threshold $\eta$. See Section~\ref{sec:experiments} and Appendix~\ref{app:exp_details} for the exact criteria.
Alternative means of ensuring invertibility could be considered, but we note that typical regularization techniques like Tikhonov regularization would not apply, due to heavily influencing $\lambda_{\text{min}}$ and hence especially Eq.~\ref{eq:hilbert_dist} that only depends on the smallest and largest eigenvalues.
\begin{table}[t]
\centering
\caption{Reconstruction error ($\downarrow$) of the test set in the target domain for image denoising.}
\label{tab:img_denoising_res}
\begin{tabular}{c|c|cc}
Method                  & Moment   & MNIST           & Fashion-MNIST    \\ \hline
Source-only             & -        & $0.094\pm0.012$ & $0.159\pm0.005$  \\ \hline
DDC                     & 1        & $0.078\pm0.001$ & $0.112\pm0.004$  \\
DCORAL                  & 2        & $0.080\pm0.003$ & $0.070\pm0.005$  \\
MECA                    & 2        & $0.077\pm0.001$ & $0.070\pm0.003$  \\
CMD                     & 1, 2     & $0.073\pm0.003$ & $0.074\pm0.002$  \\
HoMM                    & 1, 2     & $0.087\pm0.0$   & $0.076\pm0.007$  \\ \hline
CMD                     & 1, 2, 3  & $0.073\pm0.003$ & $0.071\pm0.004$  \\
HoMM                    & 1, 2, 3  & $0.092\pm0.004$ & $0.159\pm0.008$  \\ \hline
GeoAdapt-HPD (ours) & 1, 2     & $\bf{0.059\pm0.001}$ & $\bf{0.050\pm0.001}$  \\
GeoAdapt-AIRD     (ours) & 1, 2     & $0.061\pm0.001$ & $\bf{0.050\pm0.001}$

\end{tabular}%
\end{table}

\section{Experiments \& Results}
\label{sec:experiments}
We evaluate our approach on both ST and UT tasks. Note that the adaptation itself is always carried out in a fully unsupervised manner, independent of the downstream task. For ST, we follow prior work on moment-matching for UDA and consider image classification. For UT, we demonstrate the broader applicability of our method through image denoising.

\paragraph{Comparison methods.} We benchmark our method with two choices for the distance, labeled \textit{GeoAdapt-HPD}, where we use $d_H$ as the $\mathcal{L}_{\text{dist}}$, and \textit{GeoAdapt-AIRD}, where $\mathcal{L}_{\text{dist}}$ is set to $d_A$, against several representative moment-matching UDA methods: DDC~\citep{tzeng2014deepdomainconfusionmaximizing}, DCORAL~\citep{10.1007/978-3-319-49409-8_35}, MECA~\citep{morerio2018minimalentropy}, CMD~\citep{DBLP:conf/iclr/ZellingerGLNS17}, and HoMM~\citep{Chen_Fu_Chen_Jin_Cheng_Jin_Hua_2020}. Among these, only MECA employs a geometrically motivated distance (log-Euclidean) to compare source and target distributions. All methods share the same general loss in Eq.~\ref{eq:loss_factorization}, and we use the same architecture for all, including the same embedding dimensionality $n$, chosen to be the largest one for which $P_S$ is robustly invertible for the given data. We also include a \textit{Source-only} baseline trained without any adaptation.
For CMD and HoMM, which support higher-order matching, we report results using both the first two and the first three moments.

\subsection{Unsupervised Down-Stream Task: Image Denoising}
\label{sec:img_den}

\paragraph{Data \& Setup.} We evaluate image denoising on \textit{MNIST} and \textit{Fashion-MNIST}. Clean images serve as the source domain, while noisy images form the target domain. Following \citet{9008549}, we corrupt half of the images in each train/test split by adding Gaussian noise $\omega \sim N(0.4, 0.7^2)$. Moreover, the source and target domains consist of distinct, non-paired images. The goal is to map noisy target images into a latent space where reconstructions resemble clean source images. We train an autoencoder identical to that of \citet{9008549} with two-dimensional embedding layer, with mean squared error as $\mathcal{L}_{\text{task}}$ (Eq.~\ref{eq:loss_factorization}). The results are reported on the noisy target test samples, with further experimental details including the choice of the hyperparameters provided in Appendix~\ref{app:exp_img_den}. 

\paragraph{Results.} Table~\ref{tab:img_denoising_res} shows the average reconstruction error on the noisy target test samples, averaged over three runs. On both datasets, our methods consistently outperform all baselines, including CMD and HoMM with higher-order moment matching. We also observe that incorporating additional moments does not always improve performance -- evident in HoMM -- echoing findings from \citet{Chen_Fu_Chen_Jin_Cheng_Jin_Hua_2020}, where matching beyond a certain order degraded adaptation quality.

\begin{table}[]
\centering
\caption{Classification accuracy ($\uparrow$) on the target domain for the Office-31 benchmark.}
\label{tab:img_cls_res}
\resizebox{\columnwidth}{!}{%
\begin{tabular}{c|c|ccccc|c}
Method                  & Moment   & A$\rightarrow$W & D$\rightarrow$W & A$\rightarrow$D & D$\rightarrow$A & W$\rightarrow$A &  Avg    \\ \hline
Source-Only             & -        & $0.698\pm0.001$ & $0.950\pm0.001$ & $0.714\pm0.018$ & $0.597\pm0.01$  & $0.601\pm0.011$ & $0.712$ \\ \hline
DDC                     & 1        & $0.786\pm0.016$ & $\bf{0.962\pm0.002}$ & $\bf{0.846\pm0.030}$ & $0.599\pm0.016$ & $0.596\pm0.018$ & $0.758$ \\
DCORAL                  & 2        & $0.797\pm0.006$ & $0.867\pm0.01$  & $0.776\pm0.002$ & $0.604\pm0.014$ & $0.637\pm0.037$ & $0.736$ \\
MECA                    & 2        & $0.800\pm0.010$ & $\bf{0.962\pm0.003}$ & $0.776\pm0.007$ & $0.632\pm0.006$ & $0.647\pm0.008$ & $0.763$ \\
CMD                     & 1, 2     & $0.774\pm0.018$ & $0.946\pm0.003$ & $0.792\pm0.006$ & $0.557\pm0.036$ & $0.555\pm0.005$ & $0.725$ \\
HoMM                    & 1, 2     & $0.797\pm0.012$ & $0.931\pm0.004$ & $0.776\pm0.007$ & $0.580\pm0.021$ & $0.601\pm0.026$ & $0.737$ \\ \hline
CMD                     & 1, 2, 3  & $0.789\pm0.002$ & $0.953\pm0.001$ & $0.809\pm0.017$ & $0.602\pm0.018$ & $0.610\pm0.009$ & $0.753$ \\
HoMM                    & 1, 2, 3  & $0.835\pm0.019$ & $0.950\pm0.004$ & $0.814\pm0.006$ & $0.619\pm0.012$ & $0.624\pm0.022$ & $0.768$ \\ \hline
GeoAdapt-HPD (ours) & 1, 2         & $0.830\pm0.004$ & $\bf{0.962\pm0.002}$ & $0.817\pm0.006$ & $0.606\pm0.011$ & $0.624\pm0.013$ & $0.768$ \\
GeoAdapt-AIRD  (ours)    & 1, 2    & $\bf{0.846\pm0.009}$ & $0.961\pm0.003$ & $0.828\pm0.005$ & $\bf{0.647\pm0.009}$ & $\bf{0.661\pm0.010}$ & $\bf{0.789}$
\end{tabular}%
}
\end{table}

\subsection{Supervised Down-Stream Task: Image Classification}
\label{sec:img_cls}
We evaluate classification under domain shift using two standard DA benchmarks. The \textbf{Office-31} data \citep{10.1007/978-3-642-15561-1_16} contains three domains: \textit{Amazon} (A), \textit{DSLR} (D), and \textit{Webcam} (W). We construct six source–target transfer tasks by treating one domain as the source and another as the target. Following common practice, we exclude the W$\rightarrow$D task because classification accuracy on this pair remains nearly perfect even without adaptation, making it uninformative for evaluation. The \textbf{VisDA-2017} data \citep{visda2017} is designed for large-scale, challenging DA. It consists of three domains: a training domain with synthetic renderings of 3D objects, a validation domain with cropped images from Microsoft COCO~\citep{10.1007/978-3-319-10602-1_48}, and a test domain with cropped images from YouTube-BoundingBox~\citep{8100272}. We tune hyperparameters on the validation domain and report results on the test domain as the primary adaptation target.

\textbf{Backbone model.} For both benchmarks, we adopt ResNet-50~\citep{7780459} pretrained on ImageNet as the backbone. A fully connected adaptation layer is added to extract latent features, followed by a classification head whose output dimension matches the number of dataset-specific classes, similar to \citet{Chen_Fu_Chen_Jin_Cheng_Jin_Hua_2020}. The adaptation layer dimensionality is set to $42$ for Office-31 and $25$ for VisDA-2017. See Appendix~\ref{app:exp_img_cls} for full details and justification for the choices.

\paragraph{Results.} Table~\ref{tab:img_cls_res} reports accuracy on Office-31, averaged over three independent runs,
using the same hyperparameters for all tasks to demonstrate robustness of the approaches.
The final column summarizes the average performance across the five transfer setups. Overall, \textit{GeoAdapt-AIRD} is overall the best with very reliable performance, and the the next best methods (\textit{GeoAdapt-HPD} and \textit{HoMM} with 3 moments) that also use geometry-aware distances are also ahead of the rest. The D$\rightarrow$A and W$\rightarrow$A tasks are challenging for most methods, due to small source domains.

Table~\ref{tab:visda} presents results on VisDA-2017, where adaptation must succeed in an out-of-the-box deployment scenario: the target domain is unseen during hyperparameter tuning. Results are averaged over ten runs. \textit{GeoAdapt-AIRD} is again the best, followed also by the geometry-aware \textit{MECA}.

\section{Discussion}
\label{sec:disc}
\begin{figure}[t]
\centering
\includegraphics[scale = 0.24]{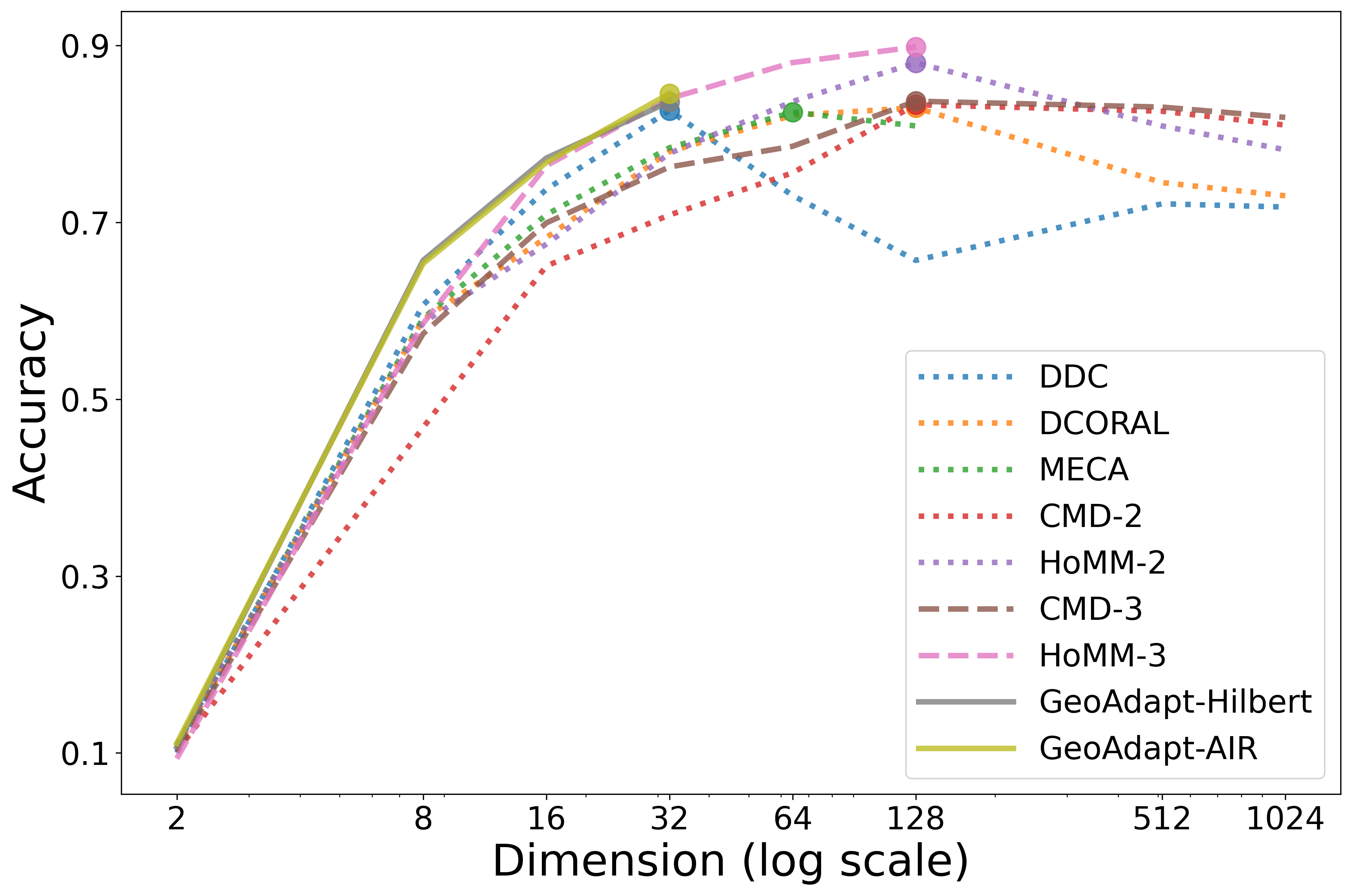}
\caption{Accuracy on the A$\rightarrow$W setup of the Office-31 dataset as a function of the embedding dimensionality (x-axis). All methods achieve the best accuracy (marked with a point) for dimensionality substantially lower than the full ResNet embedding space. Our distances are the best for the dimensionalities up to our conservative choice of maximum dimensionality where $P_S$ can be robustly inverted.}
\label{fig:dim}
\end{figure}

\paragraph{Feature dimensionality.}
We used compact embedding spaces of dimensionality in the order of tens, in contrast to most previous works using the full ResNet embeddings. While we motivated this in part by ensuring invertibility, the question of the right embedding dimensionality is more profound. Figure~\ref{fig:dim} shows the performance of the various methods on Office-31 as a function of the dimensionality $n$, revealing that it is beneficial to use a compact adaptation layer for \emph{all} baseline methods as well: Each method achieves the highest accuracy with $n \in [32, 128]$. This suggests people should consider reduced-dimensional embeddings in DA tasks more broadly, with possibility of gaining both accuracy and computational efficiency. Both of our distances are consistently the best for low-to-mid dimensionalities, and likely they could be made computable also for higher dimensionality e.g. by considering large mini-batches or covariance shrinkage methods~\citep{Ledoit2003}. We intentionally used a conservative strategy where computational issues are guaranteed to be avoided, not exploring approximations for higher dimensionalities.

\paragraph{Analysis.}
Our work also helps to understand phenomena such as the one reported in Fig.~\ref{fig:dim}. Although methods relying on the Euclidean distance between moments can be formally computed in high dimensions, they are \emph{expected} to fail at some point. This occurs because when $b \ll n$, the covariances are rank-deficient and lie near the boundary of the SPD manifold. In this region, the curvature is more pronounced, and the Euclidean distance becomes especially misleading compared to the true geodesic distance within the manifold of SPD matrices.~\citep{Pennec2006, pmlr-v221-nielsen23b, 10.1007/978-3-319-10605-2_2}. In other words, by merely inspecting the problem from the perspective of the appropriate embedding space and metric, we can explain also failure modes of classical methods.

\paragraph{Empirical performance.}
We showed improvement over the leading moment matching comparison methods in targeted experiments, designed to isolate the effect of the distance metric. In terms of absolute performance, the current-state-of-the art (e.g. \citet{9578777}) report clearly higher accuracies. This is because of substantially stronger backbones (e.g. ResNet-101 or transformers), adaptation of the entire network rather than the final layers only, and various advanced techniques like pseudo-labeling on the target domain and explicit modeling of class-discriminative structures \citep{Luo_Ren_Ge_Huang_Yu_2020,Dai_2020_ACCV,Chen_Chen_Jiang_Jin_2019}. These enhancements are orthogonal to our contribution: our distance can be plugged into any method that uses the loss factorization of Eq.~\ref{eq:loss_factorization}. We leave the evaluation of such methods to future work. 

\begin{table}[t]
\centering
\caption{Classification accuracy ($\uparrow$) on the target domain for the VisDA-2017 benchmark.}
\label{tab:visda}
\begin{tabular}{c|c|c}
Method                   & Moment   & Accuracy       \\ \hline
Source-only              & -        & $0.345\pm0.021$  \\ \hline
DDC                      & 1        & $0.526\pm0.016$  \\
DCORAL                   & 2        & $0.700\pm0.012$  \\
MECA                     & 2        & $0.736\pm0.014$  \\
CMD                      & 1, 2     & $0.634\pm0.038$  \\
HoMM                     & 1, 2     & $0.717\pm0.007$  \\ \hline
CMD                      & 1, 2, 3  & $0.733\pm0.046$  \\
HoMM                     & 1, 2, 3  & $0.705\pm0.028$  \\ \hline
GeoAdapt-HPD (ours)  & 1, 2     & $0.715\pm0.022$  \\
GeoAdapt-AIRD  (ours)     & 1, 2     & $\bf{0.748\pm0.021}$  \\
\end{tabular}%
\end{table}

\section{Conclusion}
\label{sec:conclusion}
We improve moment matching methods for unsupervised domain adaptation by better accounting for the intrinsic non-Euclidean geometry of the moments. We embed the first- and second-order moments of the source and target probability distributions into the SPD matrix manifold, measuring the domain discrepancy on this manifold. We explored two complementary distances: the affine-invariant Riemannian distance and the Hilbert projective distance, and demonstrated that these geometry-aware distances improve the performance on image benchmarks. For the latter we have a formal upper bound on the generalization error, but the former is generally more accurate. We also showed that surprisingly low-dimensional feature spaces are good for adaptation, not just for our metrics but in general.
Our experiments focused specifically on quantifying the effect of the geometric distance as a plug-in replacement for the domain discrepancy loss. The improvement is expected to translate to the broad range of more DA methods that share the same general form.

\section*{Acknowledgments}
This project has received funding from the European Union – NextGenerationEU instrument and is funded by the Research Council of Finland under grant number 353411. We additionally acknowledge support from the  Research Council of Finland Flagship programme: Finnish Center for Artificial Intelligence FCAI, and grants 345811, 363317, 348952 and 369502.
The authors wish to acknowledge CSC - IT Center for Science, Finland, for computational resources. The authors acknowledge the research environment provided by ELLIS Institute Finland.

\newpage

\bibliography{neurips_2025}

\begin{thebibliography}{80}
\providecommand{\natexlab}[1]{#1}
\providecommand{\url}[1]{\texttt{#1}}
\expandafter\ifx\csname urlstyle\endcsname\relax
  \providecommand{\doi}[1]{doi: #1}\else
  \providecommand{\doi}{doi: \begingroup \urlstyle{rm}\Url}\fi

\bibitem[Absil et~al.(2008)Absil, Mahony, and Sepulchre]{absil:2008}
P.-A. Absil, R.~Mahony, and R.~Sepulchre.
\newblock \emph{Optimization Algorithms on Matrix Manifolds}.
\newblock Princeton University Press, 2008.

\bibitem[Amari(1998)]{amari:1998}
Shun-ichi Amari.
\newblock Natural gradient works efficiently in learning.
\newblock \emph{Neural Computation}, 10\penalty0 (2):\penalty0 251--276, 1998.

\bibitem[Amari \& Nagaoka(2000)Amari and Nagaoka]{amari2000methods}
Shun-Ichi Amari and Hiroshi. Nagaoka.
\newblock \emph{Methods of Information Geometry}.
\newblock Translations of mathematical monographs. 2000.

\bibitem[Arsigny et~al.(2007)Arsigny, Fillard, Pennec, and Ayache]{doi:10.1137/050637996}
Vincent Arsigny, Pierre Fillard, Xavier Pennec, and Nicholas Ayache.
\newblock Geometric means in a novel vector space structure on symmetric positive‐definite matrices.
\newblock \emph{SIAM Journal on Matrix Analysis and Applications}, 29\penalty0 (1):\penalty0 328--347, 2007.

\bibitem[Balaji et~al.(2019)Balaji, Chellappa, and Feizi]{9008549}
Yogesh Balaji, Rama Chellappa, and Soheil Feizi.
\newblock Normalized wasserstein for mixture distributions with applications in adversarial learning and domain adaptation.
\newblock In \emph{2019 IEEE/CVF International Conference on Computer Vision (ICCV)}, pp.\  6499--6507, 2019.

\bibitem[Ben-David et~al.(2010)Ben-David, Blitzer, Crammer, Kulesza, Pereira, and Vaughan]{Ben-David2010}
Shai Ben-David, John Blitzer, Koby Crammer, Alex Kulesza, Fernando Pereira, and Jennifer~Wortman Vaughan.
\newblock A theory of learning from different domains.
\newblock \emph{Machine Learning}, 79\penalty0 (1):\penalty0 151--175, May 2010.

\bibitem[Bernstein(2009)]{bernstein2009matrix}
D.S. Bernstein.
\newblock \emph{Matrix Mathematics: Theory, Facts, and Formulas - Second Edition}.
\newblock Princeton University Press, 2009.

\bibitem[Bhatia(2007)]{Bhatia+2007}
Rajendra Bhatia.
\newblock \emph{Positive Definite Matrices}.
\newblock Princeton, 2007.

\bibitem[Bronstein et~al.(2017)Bronstein, Bruna, LeCun, Szlam, and Vandergheynst]{bronstein:2017}
Michael~M. Bronstein, Joan Bruna, Yann LeCun, Arthur Szlam, and Pierre Vandergheynst.
\newblock Geometric deep learning: Going beyond euclidean data.
\newblock \emph{IEEE Signal Processing Magazine}, 34\penalty0 (4):\penalty0 18--42, 2017.

\bibitem[Brooks et~al.(2019)Brooks, Schwander, Barbaresco, Schneider, and Cord]{brooks:2019}
Daniel Brooks, Olivier Schwander, Fr{\'e}d{\'e}ric Barbaresco, Jean-Yves Schneider, and Matthieu Cord.
\newblock Riemannian batch normalization for {SPD} neural networks.
\newblock In \emph{Advances in Neural Information Processing Systems}, pp.\  15489--15500, 2019.

\bibitem[Calin \& Udrişte(2014)Calin and Udrişte]{calin:2014}
Ovidiu Calin and Constantin Udrişte.
\newblock \emph{Geometric Modeling in Probability and Statistics}.
\newblock Springer International Publishing, 1 edition, 2014.

\bibitem[Calvo \& Oller(1990)Calvo and Oller]{CALVO1990223}
Miquel Calvo and Josep~M. Oller.
\newblock A distance between multivariate normal distributions based in an embedding into the siegel group.
\newblock \emph{Journal of Multivariate Analysis}, 35\penalty0 (2):\penalty0 223--242, 1990.

\bibitem[Chen et~al.(2019)Chen, Chen, Jiang, and Jin]{Chen_Chen_Jiang_Jin_2019}
Chao Chen, Zhihong Chen, Boyuan Jiang, and Xinyu Jin.
\newblock Joint domain alignment and discriminative feature learning for unsupervised deep domain adaptation.
\newblock \emph{Proceedings of the AAAI Conference on Artificial Intelligence}, 33\penalty0 (01):\penalty0 3296--3303, Jul. 2019.

\bibitem[Chen et~al.(2020)Chen, Fu, Chen, Jin, Cheng, Jin, and Hua]{Chen_Fu_Chen_Jin_Cheng_Jin_Hua_2020}
Chao Chen, Zhihang Fu, Zhihong Chen, Sheng Jin, Zhaowei Cheng, Xinyu Jin, and Xian-sheng Hua.
\newblock Homm: Higher-order moment matching for unsupervised domain adaptation.
\newblock \emph{Proceedings of the AAAI Conference on Artificial Intelligence}, 34\penalty0 (04):\penalty0 3422--3429, Apr. 2020.

\bibitem[Chen et~al.(2022)Chen, Wei, Jin, Chen, Zheng, Chen, and Jin]{NEURIPS2022_61aa5576}
Lin Chen, Zhixiang Wei, Xin Jin, Huaian Chen, Miao Zheng, Kai Chen, and Yi~Jin.
\newblock Deliberated domain bridging for domain adaptive semantic segmentation.
\newblock In \emph{Advances in Neural Information Processing Systems}, volume~35, pp.\  15105--15118, 2022.

\bibitem[Cohen \& Fausti(2024)Cohen and Fausti]{cohen2024hyperboliccontractivityhilbertmetric}
Samuel~N. Cohen and Eliana Fausti.
\newblock Hyperbolic contractivity and the {H}ilbert metric on probability measures.
\newblock \emph{ArXiv:2309.02413}, 2024.

\bibitem[Dai et~al.(2020)Dai, Cheng, Zhang, Gan, Liu, and Carin]{Dai_2020_ACCV}
Shuyang Dai, Yu~Cheng, Yizhe Zhang, Zhe Gan, Jingjing Liu, and Lawrence Carin.
\newblock Contrastively smoothed class alignment for unsupervised domain adaptation.
\newblock In \emph{Proceedings of the Asian Conference on Computer Vision (ACCV)}, November 2020.

\bibitem[Do~Carmo(1992)]{docarmo:1992}
Manfredo~P Do~Carmo.
\newblock \emph{Riemannian Geometry}.
\newblock Mathematics. Theory \& applications. Birkhäuser, 1992.

\bibitem[Do~Carmo(2017)]{docarmo:2016}
Manfredo~P. Do~Carmo.
\newblock \emph{Differential Geometry of Curves and Surfaces}.
\newblock Dover Publications, 2nd edition edition, 2017.

\bibitem[Ganin \& Lempitsky(2015)Ganin and Lempitsky]{pmlr-v37-ganin15}
Yaroslav Ganin and Victor Lempitsky.
\newblock Unsupervised domain adaptation by backpropagation.
\newblock In \emph{Proceedings of the 32nd International Conference on Machine Learning}, volume~37 of \emph{Proceedings of Machine Learning Research}, pp.\  1180--1189, 07--09 Jul 2015.

\bibitem[Golub \& van Loan(2013)Golub and van Loan]{golub13}
Gene~H. Golub and Charles~F. van Loan.
\newblock \emph{Matrix Computations}.
\newblock Johns Hopkins University Pres, fourth edition, 2013.

\bibitem[Harandi et~al.(2014)Harandi, Salzmann, and Hartley]{10.1007/978-3-319-10605-2_2}
Mehrtash~T. Harandi, Mathieu Salzmann, and Richard Hartley.
\newblock From manifold to manifold: Geometry-aware dimensionality reduction for spd matrices.
\newblock In \emph{Computer Vision -- ECCV 2014}, pp.\  17--32, 2014.

\bibitem[He et~al.(2016)He, Zhang, Ren, and Sun]{7780459}
Kaiming He, Xiangyu Zhang, Shaoqing Ren, and Jian Sun.
\newblock Deep residual learning for image recognition.
\newblock In \emph{2016 IEEE Conference on Computer Vision and Pattern Recognition (CVPR)}, pp.\  770--778, 2016.
\newblock \doi{10.1109/CVPR.2016.90}.

\bibitem[Huang \& Van~Gool(2017)Huang and Van~Gool]{Huang_Van_Gool_2017}
Zhiwu Huang and Luc Van~Gool.
\newblock A riemannian network for spd matrix learning.
\newblock \emph{Proceedings of the AAAI Conference on Artificial Intelligence}, 31\penalty0 (1), Feb. 2017.

\bibitem[Jayasumana et~al.(2013)Jayasumana, Hartley, Salzmann, Li, and Harandi]{Jayasumana_2013_CVPR}
Sadeep Jayasumana, Richard Hartley, Mathieu Salzmann, Hongdong Li, and Mehrtash Harandi.
\newblock Kernel methods on the riemannian manifold of symmetric positive definite matrices.
\newblock In \emph{Proceedings of the IEEE Conference on Computer Vision and Pattern Recognition (CVPR)}, June 2013.

\bibitem[Kalischek et~al.(2021)Kalischek, Wegner, and Schindler]{Kalischek_2021_CVPR}
Nikolai Kalischek, Jan~D. Wegner, and Konrad Schindler.
\newblock In the light of feature distributions: Moment matching for neural style transfer.
\newblock In \emph{Proceedings of the IEEE/CVF Conference on Computer Vision and Pattern Recognition (CVPR)}, pp.\  9382--9391, June 2021.

\bibitem[Kang et~al.(2019)Kang, Jiang, Yang, and Hauptmann]{8954037}
Guoliang Kang, Lu~Jiang, Yi~Yang, and Alexander~G. Hauptmann.
\newblock Contrastive adaptation network for unsupervised domain adaptation.
\newblock In \emph{2019 IEEE/CVF Conference on Computer Vision and Pattern Recognition (CVPR)}, pp.\  4888--4897, 2019.

\bibitem[Kass \& Vos(1997)Kass and Vos]{kass::1997}
Robert~E. Kass and Paul~W. Vos.
\newblock \emph{Geometrical Foundations of Asymptotic Inference}.
\newblock Probability and Statistics 125. Wiley-Interscience, 1997.

\bibitem[Kingma \& Ba(2014)Kingma and Ba]{kingma2014adam}
Diederik~P Kingma and Jimmy Ba.
\newblock Adam: A method for stochastic optimization.
\newblock \emph{arXiv preprint arXiv:1412.6980}, 2014.

\bibitem[Knyazev(2001)]{doi:10.1137/S1064827500366124}
Andrew~V. Knyazev.
\newblock Toward the optimal preconditioned eigensolver: Locally optimal block preconditioned conjugate gradient method.
\newblock \emph{SIAM Journal on Scientific Computing}, 23\penalty0 (2):\penalty0 517--541, 2001.

\bibitem[Kobler et~al.(2022)Kobler, Hirayama, Zhao, and Kawanabe]{NEURIPS2022_28ef7ee7}
Reinmar Kobler, Jun-ichiro Hirayama, Qibin Zhao, and Motoaki Kawanabe.
\newblock Spd domain-specific batch normalization to crack interpretable unsupervised domain adaptation in eeg.
\newblock In \emph{Advances in Neural Information Processing Systems}, volume~35, pp.\  6219--6235, 2022.

\bibitem[Koh et~al.(2021)Koh, Sagawa, Marklund, Xie, Zhang, Balsubramani, Hu, Yasunaga, Phillips, Gao, Lee, David, Stavness, Guo, Earnshaw, Haque, Beery, Leskovec, Kundaje, Pierson, Levine, Finn, and Liang]{pmlr-v139-koh21a}
Pang~Wei Koh, Shiori Sagawa, Henrik Marklund, Sang~Michael Xie, Marvin Zhang, Akshay Balsubramani, Weihua Hu, Michihiro Yasunaga, Richard~Lanas Phillips, Irena Gao, Tony Lee, Etienne David, Ian Stavness, Wei Guo, Berton Earnshaw, Imran Haque, Sara~M Beery, Jure Leskovec, Anshul Kundaje, Emma Pierson, Sergey Levine, Chelsea Finn, and Percy Liang.
\newblock Wilds: A benchmark of in-the-wild distribution shifts.
\newblock In \emph{Proceedings of the 38th International Conference on Machine Learning}, volume 139 of \emph{Proceedings of Machine Learning Research}, pp.\  5637--5664, 18--24 Jul 2021.

\bibitem[Kurowicka \& Cooke(2003)Kurowicka and Cooke]{kurowicka:2003}
Dorota Kurowicka and Roger Cooke.
\newblock A parameterization of positive definite matrices in terms of partial correlation vines.
\newblock \emph{Linear Algebra and its Applications}, 372:\penalty0 225--251, 2003.

\bibitem[Ledoit \& Wolf(2003)Ledoit and Wolf]{Ledoit2003}
Olivier Ledoit and Michael Wolf.
\newblock Honey, i shrunk the sample covariance matrix.
\newblock Technical Report 691, UPF Economics and Business, June 2003.

\bibitem[Levin \& Peres(2017)Levin and Peres]{levin2017markov}
D.A. Levin and Y.~Peres.
\newblock \emph{Markov Chains and Mixing Times}.
\newblock MBK. American Mathematical Society, 2017.

\bibitem[Lin et~al.(2014)Lin, Maire, Belongie, Hays, Perona, Ramanan, Doll{\'a}r, and Zitnick]{10.1007/978-3-319-10602-1_48}
Tsung-Yi Lin, Michael Maire, Serge Belongie, James Hays, Pietro Perona, Deva Ramanan, Piotr Doll{\'a}r, and C.~Lawrence Zitnick.
\newblock Microsoft coco: Common objects in context.
\newblock In \emph{Computer Vision -- ECCV 2014}, pp.\  740--755, 2014.

\bibitem[Long et~al.(2015)Long, Cao, Wang, and Jordan]{pmlr-v37-long15}
Mingsheng Long, Yue Cao, Jianmin Wang, and Michael Jordan.
\newblock Learning transferable features with deep adaptation networks.
\newblock In \emph{Proceedings of the 32nd International Conference on Machine Learning}, volume~37 of \emph{Proceedings of Machine Learning Research}, pp.\  97--105, 07--09 Jul 2015.

\bibitem[Long et~al.(2016)Long, Zhu, Wang, and Jordan]{NIPS2016_ac627ab1}
Mingsheng Long, Han Zhu, Jianmin Wang, and Michael~I Jordan.
\newblock Unsupervised domain adaptation with residual transfer networks.
\newblock In \emph{Advances in Neural Information Processing Systems}, volume~29, 2016.

\bibitem[Long et~al.(2017)Long, Zhu, Wang, and Jordan]{pmlr-v70-long17a}
Mingsheng Long, Han Zhu, Jianmin Wang, and Michael~I. Jordan.
\newblock Deep transfer learning with joint adaptation networks.
\newblock In \emph{Proceedings of the 34th International Conference on Machine Learning}, volume~70 of \emph{Proceedings of Machine Learning Research}, pp.\  2208--2217, 06--11 Aug 2017.

\bibitem[Luo et~al.(2020)Luo, Ren, Ge, Huang, and Yu]{Luo_Ren_Ge_Huang_Yu_2020}
You-Wei Luo, Chuan-Xian Ren, Pengfei Ge, Ke-Kun Huang, and Yu-Feng Yu.
\newblock Unsupervised domain adaptation via discriminative manifold embedding and alignment.
\newblock \emph{Proceedings of the AAAI Conference on Artificial Intelligence}, 34\penalty0 (04):\penalty0 5029--5036, Apr. 2020.

\bibitem[Martens(2020)]{martens:2020}
James Martens.
\newblock New insights and perspectives on the natural gradient method.
\newblock \emph{Journal of Machine Learning Research}, 21\penalty0 (146):\penalty0 1--76, 2020.

\bibitem[Minh et~al.(2014)Minh, San~Biagio, and Murino]{NIPS2014_3000e56b}
H\`{a}~Quang Minh, Marco San~Biagio, and Vittorio Murino.
\newblock Log-hilbert-schmidt metric between positive definite operators on hilbert spaces.
\newblock In \emph{Advances in Neural Information Processing Systems}, volume~27, 2014.

\bibitem[Miolane et~al.(2020)Miolane, Lebrigand, Mathe, Pennec, et~al.]{miolane:2020}
Nina Miolane, Claire Lebrigand, Johan Mathe, Xavier Pennec, et~al.
\newblock Geomstats: A python package for riemannian geometry in machine learning.
\newblock \emph{Journal of Machine Learning Research}, 21\penalty0 (223):\penalty0 1--9, 2020.

\bibitem[Moreno-Torres et~al.(2012)Moreno-Torres, Raeder, Alaiz-Rodríguez, Chawla, and Herrera]{MORENOTORRES2012521}
Jose~G. Moreno-Torres, Troy Raeder, Rocío Alaiz-Rodríguez, Nitesh~V. Chawla, and Francisco Herrera.
\newblock A unifying view on dataset shift in classification.
\newblock \emph{Pattern Recognition}, 45\penalty0 (1):\penalty0 521--530, 2012.

\bibitem[Morerio et~al.(2018)Morerio, Cavazza, and Murino]{morerio2018minimalentropy}
Pietro Morerio, Jacopo Cavazza, and Vittorio Murino.
\newblock Minimal-entropy correlation alignment for unsupervised deep domain adaptation.
\newblock In \emph{International Conference on Learning Representations}, 2018.

\bibitem[Na et~al.(2021)Na, Jung, Chang, and Hwang]{9578777}
Jaemin Na, Heechul Jung, Hyung~Jin Chang, and Wonjun Hwang.
\newblock Fixbi: Bridging domain spaces for unsupervised domain adaptation.
\newblock In \emph{2021 IEEE/CVF Conference on Computer Vision and Pattern Recognition (CVPR)}, pp.\  1094--1103, 2021.

\bibitem[Nickel \& Kiela(2017)Nickel and Kiela]{nickel:2017}
Maximilian Nickel and Douwe Kiela.
\newblock Poincar{\'e} embeddings for learning hierarchical representations.
\newblock In \emph{Advances in Neural Information Processing Systems}, pp.\  6338--6347, 2017.

\bibitem[Nielsen(2023{\natexlab{a}})]{e25040654}
Frank Nielsen.
\newblock A simple approximation method for the {F}isher–{R}ao distance between multivariate normal distributions.
\newblock \emph{Entropy}, 25\penalty0 (4), 2023{\natexlab{a}}.

\bibitem[Nielsen(2023{\natexlab{b}})]{pmlr-v221-nielsen23b}
Frank Nielsen.
\newblock Fisher-{R}ao and pullback {H}ilbert cone distances on the multivariate {G}aussian manifold with applications to simplification and quantization of mixtures.
\newblock In \emph{Proceedings of 2nd Annual Workshop on Topology, Algebra, and Geometry in Machine Learning (TAG-ML)}, volume 221 of \emph{Proceedings of Machine Learning Research}, pp.\  488--504, 28 Jul 2023{\natexlab{b}}.

\bibitem[Peng et~al.(2017)Peng, Usman, Kaushik, Hoffman, Wang, and Saenko]{visda2017}
Xingchao Peng, Ben Usman, Neela Kaushik, Judy Hoffman, Dequan Wang, and Kate Saenko.
\newblock Visda: The visual domain adaptation challenge.
\newblock \emph{ArXiv:1710.06924}, 2017.

\bibitem[Pennec et~al.(2006)Pennec, Fillard, and Ayache]{Pennec2006}
Xavier Pennec, Pierre Fillard, and Nicholas Ayache.
\newblock {A Riemannian Framework for Tensor Computing}.
\newblock \emph{{International Journal of Computer Vision}}, 66\penalty0 (1):\penalty0 41--66, 2006.

\bibitem[Petersen(2016)]{petersen:2016}
Peter Petersen.
\newblock \emph{Riemannian Geometry}.
\newblock Springer, 3rd edition, 2016.

\bibitem[Pinele et~al.(2020)Pinele, Strapasson, and Costa]{pinele:2020}
Julianna Pinele, João~E. Strapasson, and Sueli I.~R. Costa.
\newblock The fisher–rao distance between multivariate normal distributions: Special cases, bounds and applications.
\newblock \emph{Entropy}, 22\penalty0 (4), 2020.

\bibitem[Quionero-Candela et~al.(2009)Quionero-Candela, Sugiyama, Schwaighofer, and Lawrence]{10.5555/1462129}
Joaquin Quionero-Candela, Masashi Sugiyama, Anton Schwaighofer, and Neil~D. Lawrence.
\newblock \emph{Dataset Shift in Machine Learning}.
\newblock The MIT Press, 2009.

\bibitem[Rangwani et~al.(2022)Rangwani, Aithal, Mishra, Jain, and Radhakrishnan]{pmlr-v162-rangwani22a}
Harsh Rangwani, Sumukh~K Aithal, Mayank Mishra, Arihant Jain, and Venkatesh~Babu Radhakrishnan.
\newblock A closer look at smoothness in domain adversarial training.
\newblock In \emph{Proceedings of the 39th International Conference on Machine Learning}, volume 162 of \emph{Proceedings of Machine Learning Research}, pp.\  18378--18399, 17--23 Jul 2022.

\bibitem[Real et~al.(2017)Real, Shlens, Mazzocchi, Pan, and Vanhoucke]{8100272}
Esteban Real, Jonathon Shlens, Stefano Mazzocchi, Xin Pan, and Vincent Vanhoucke.
\newblock { YouTube-BoundingBoxes: A Large High-Precision Human-Annotated Data Set for Object Detection in Video }.
\newblock In \emph{2017 IEEE Conference on Computer Vision and Pattern Recognition (CVPR)}, pp.\  7464--7473, July 2017.

\bibitem[Recht et~al.(2019)Recht, Roelofs, Schmidt, and Shankar]{pmlr-v97-recht19a}
Benjamin Recht, Rebecca Roelofs, Ludwig Schmidt, and Vaishaal Shankar.
\newblock Do {I}mage{N}et classifiers generalize to {I}mage{N}et?
\newblock In \emph{Proceedings of the 36th International Conference on Machine Learning}, volume~97 of \emph{Proceedings of Machine Learning Research}, pp.\  5389--5400, 09--15 Jun 2019.

\bibitem[Rousseeuw \& Molenberghs(1994)Rousseeuw and Molenberghs]{peter:1994}
Peter~J. Rousseeuw and Geert Molenberghs.
\newblock The shape of correlation matrices.
\newblock \emph{The American Statistician}, 48\penalty0 (4):\penalty0 276--279, 1994.

\bibitem[Saenko et~al.(2010)Saenko, Kulis, Fritz, and Darrell]{10.1007/978-3-642-15561-1_16}
Kate Saenko, Brian Kulis, Mario Fritz, and Trevor Darrell.
\newblock Adapting visual category models to new domains.
\newblock In \emph{Computer Vision -- ECCV 2010}, pp.\  213--226, 2010.

\bibitem[Sahoo et~al.(2021)Sahoo, Shah, Panda, Saenko, and Das]{NEURIPS2021_c47e9374}
Aadarsh Sahoo, Rutav Shah, Rameswar Panda, Kate Saenko, and Abir Das.
\newblock Contrast and mix: Temporal contrastive video domain adaptation with background mixing.
\newblock In \emph{Advances in Neural Information Processing Systems}, volume~34, pp.\  23386--23400, 2021.

\bibitem[Salimans et~al.(2024)Salimans, Mensink, Heek, and Hoogeboom]{NEURIPS2024_3f66d5cd}
Tim Salimans, Thomas Mensink, Jonathan Heek, and Emiel Hoogeboom.
\newblock Multistep distillation of diffusion models via moment matching.
\newblock In \emph{Advances in Neural Information Processing Systems}, volume~37, pp.\  36046--36070, 2024.

\bibitem[Shimodaira(2000)]{SHIMODAIRA2000227}
Hidetoshi Shimodaira.
\newblock Improving predictive inference under covariate shift by weighting the log-likelihood function.
\newblock \emph{Journal of Statistical Planning and Inference}, 90\penalty0 (2):\penalty0 227--244, 2000.

\bibitem[Skovgaard(1984)]{skovgaard:1984}
Lene~Theil Skovgaard.
\newblock A riemannian geometry of the multivariate normal model.
\newblock \emph{Scandinavian Journal of Statistics}, 11\penalty0 (4):\penalty0 211--223, 1984.

\bibitem[Sugiyama et~al.(2007)Sugiyama, Krauledat, and M{{\"u}}ller]{JMLR:v8:sugiyama07a}
Masashi Sugiyama, Matthias Krauledat, and Klaus-Robert M{{\"u}}ller.
\newblock Covariate shift adaptation by importance weighted cross validation.
\newblock \emph{Journal of Machine Learning Research}, 8\penalty0 (35):\penalty0 985--1005, 2007.

\bibitem[Sun \& Saenko(2016)Sun and Saenko]{10.1007/978-3-319-49409-8_35}
Baochen Sun and Kate Saenko.
\newblock Deep coral: Correlation alignment for deep domain adaptation.
\newblock In \emph{Computer Vision -- ECCV 2016 Workshops}, pp.\  443--450, 2016.

\bibitem[Sun et~al.(2016)Sun, Feng, and Saenko]{Sun_Feng_Saenko_2016}
Baochen Sun, Jiashi Feng, and Kate Saenko.
\newblock Return of frustratingly easy domain adaptation.
\newblock \emph{Proceedings of the AAAI Conference on Artificial Intelligence}, 30\penalty0 (1), Mar. 2016.

\bibitem[Torralba \& Efros(2011)Torralba and Efros]{5995347}
Antonio Torralba and Alexei~A. Efros.
\newblock Unbiased look at dataset bias.
\newblock In \emph{CVPR}, pp.\  1521--1528, 2011.

\bibitem[Tzeng et~al.(2014)Tzeng, Hoffman, Zhang, Saenko, and Darrell]{tzeng2014deepdomainconfusionmaximizing}
Eric Tzeng, Judy Hoffman, Ning Zhang, Kate Saenko, and Trevor Darrell.
\newblock Deep domain confusion: Maximizing for domain invariance, 2014.

\bibitem[Tzeng et~al.(2017)Tzeng, Hoffman, Saenko, and Darrell]{8099799}
Eric Tzeng, Judy Hoffman, Kate Saenko, and Trevor Darrell.
\newblock Adversarial discriminative domain adaptation.
\newblock In \emph{2017 IEEE Conference on Computer Vision and Pattern Recognition (CVPR)}, pp.\  2962--2971, 2017.

\bibitem[Wang et~al.(2023)Wang, Li, Ding, Nie, Chen, Dong, and Wang]{9478936}
Wei Wang, Haojie Li, Zhengming Ding, Feiping Nie, Junyang Chen, Xiao Dong, and Zhihui Wang.
\newblock Rethinking maximum mean discrepancy for visual domain adaptation.
\newblock \emph{IEEE Transactions on Neural Networks and Learning Systems}, 34\penalty0 (1):\penalty0 264--277, 2023.

\bibitem[Xiao et~al.(2023)Xiao, Wang, Jin, Feng, Chen, Huang, and Zhao]{NEURIPS2023_754e80f9}
Zhiqing Xiao, Haobo Wang, Ying Jin, Lei Feng, Gang Chen, Fei Huang, and Junbo Zhao.
\newblock Spa: A graph spectral alignment perspective for domain adaptation.
\newblock In \emph{Advances in Neural Information Processing Systems}, volume~36, pp.\  37252--37272, 2023.

\bibitem[Yan et~al.(2017)Yan, Ding, Li, Wang, Xu, and Zuo]{Yan_2017_CVPR}
Hongliang Yan, Yukang Ding, Peihua Li, Qilong Wang, Yong Xu, and Wangmeng Zuo.
\newblock Mind the class weight bias: Weighted maximum mean discrepancy for unsupervised domain adaptation.
\newblock In \emph{Proceedings of the IEEE Conference on Computer Vision and Pattern Recognition (CVPR)}, July 2017.

\bibitem[Zellinger et~al.(2017)Zellinger, Grubinger, Lughofer, Natschl{\"{a}}ger, and Saminger{-}Platz]{DBLP:conf/iclr/ZellingerGLNS17}
Werner Zellinger, Thomas Grubinger, Edwin Lughofer, Thomas Natschl{\"{a}}ger, and Susanne Saminger{-}Platz.
\newblock Central moment discrepancy {(CMD)} for domain-invariant representation learning.
\newblock In \emph{5th International Conference on Learning Representations, {ICLR}}, 2017.

\bibitem[Zellinger et~al.(2019)Zellinger, Moser, Grubinger, Lughofer, Natschläger, and Saminger-Platz]{ZELLINGER2019174}
Werner Zellinger, Bernhard~A. Moser, Thomas Grubinger, Edwin Lughofer, Thomas Natschläger, and Susanne Saminger-Platz.
\newblock Robust unsupervised domain adaptation for neural networks via moment alignment.
\newblock \emph{Information Sciences}, 483:\penalty0 174--191, 2019.

\bibitem[Zhang \& Davison(2021)Zhang and Davison]{9522884}
Youshan Zhang and Brian~D. Davison.
\newblock Deep spherical manifold gaussian kernel for unsupervised domain adaptation.
\newblock In \emph{2021 IEEE/CVF Conference on Computer Vision and Pattern Recognition Workshops (CVPRW)}, pp.\  4438--4447, 2021.

\bibitem[Zhang et~al.(2018)Zhang, Wang, Cai, and Song]{8434290}
Yun Zhang, Nianbin Wang, Shaobin Cai, and Lei Song.
\newblock Unsupervised domain adaptation by mapped correlation alignment.
\newblock \emph{IEEE Access}, 6:\penalty0 44698--44706, 2018.

\bibitem[Zhao et~al.(2019)Zhao, Combes, Zhang, and Gordon]{pmlr-v97-zhao19a}
Han Zhao, Remi Tachet~Des Combes, Kun Zhang, and Geoffrey Gordon.
\newblock On learning invariant representations for domain adaptation.
\newblock In \emph{Proceedings of the 36th International Conference on Machine Learning}, volume~97 of \emph{Proceedings of Machine Learning Research}, pp.\  7523--7532, 09--15 Jun 2019.

\bibitem[Zhao et~al.(2021)Zhao, wang, and Cai]{NEURIPS2021_4f284803}
Yin Zhao, minquan wang, and Longjun Cai.
\newblock Reducing the covariate shift by mirror samples in cross domain alignment.
\newblock In \emph{Advances in Neural Information Processing Systems}, volume~34, pp.\  9546--9558, 2021.

\bibitem[Zhou et~al.(2025)Zhou, Ermon, and Song]{zhou2025inductive}
Linqi Zhou, Stefano Ermon, and Jiaming Song.
\newblock Inductive moment matching.
\newblock \emph{arXiv preprint arXiv:2503.07565}, 2025.

\bibitem[Zhu et~al.(2019)Zhu, Zhuang, Wang, Chen, Shi, Wu, and He]{ZHU2019214}
Yongchun Zhu, Fuzhen Zhuang, Jindong Wang, Jingwu Chen, Zhiping Shi, Wenjuan Wu, and Qing He.
\newblock Multi-representation adaptation network for cross-domain image classification.
\newblock \emph{Neural Networks}, 119:\penalty0 214--221, 2019.

\end{thebibliography}
\bibliographystyle{unsrtnat}

\newpage

\appendix

\section{Theoretical Guarantee with a new bound}
\label{proof:theorem}
We begin by defining the specific divergence measure which will later on extend the existing upper bound on the expected error for the target domain.
\begin{definition} [$\tilde{\mathcal{H}}$-divergence] \citep{pmlr-v97-zhao19a} Let $\mathcal{H} \subseteq [0, 1]^{\mathcal{X}}$ be a hypothesis class. The discrepancy hypothesis class, $\mathcal{\tilde{H}}$, is defined as 
\begin{equation*}
    \mathcal{\tilde{H}} := \{sgn(|h(x) - h'(x)| - t) | h, h' \in \mathcal{H}, t \in [0, 1]\}.
\end{equation*}
The discrepancy divergence between two distributions $\mathbb{P}$ and $\mathbb{P}'$ is the $\tilde{\mathcal{H}}$-divergence with respect to this class
\begin{equation*}
    d_{\tilde{\mathcal{H}}}(\mathbb{P}, \mathbb{P}') := 2\sup_{A\in A_{\tilde{\mathcal{H}}}} |\mathbb{P}(A) - \mathbb{P}'(A)|
\end{equation*}
where $A_{\mathcal{\tilde{H}}}$ is the set of supports of hypotheses in $\tilde{\mathcal{H}}$ and $\mathbb{P}(A) = \int_A \mathrm{d}\mathbb{P}$ and $\mathbb{P}'(A) = \int_A \mathrm{d}\mathbb{P}'$.
\label{def:H-div}
\end{definition}
With this in place, we now state the theoretical result that provides an upper bound on the generalization error.
\begin{theorem} \citep{pmlr-v97-zhao19a} Let $\mathcal{H} \subseteq [0, 1]^{\mathcal{X}}$ be a hypothesis class, $\mathbb{P}_S$ and $\mathbb{P}_T$ be the distributions of covariates in the input space for the source and target domains respectively. For any $h \in \mathcal{H}$, the expected error on the target domain, $\varepsilon_t(h)$, is bounded by
\begin{equation*}
    \varepsilon_T(h) \leq \varepsilon_S(h) + d_{\tilde{\mathcal{H}}}(\mathbb{P}_S, \mathbb{P}_T) + \gamma
\end{equation*}
where $\varepsilon_S$ is the expected source error and $\gamma$ measures the inherent shift between the optimal source and target labeling functions.
\label{thrm:old_upper_bound}
\end{theorem}

Our proposed loss $\mathcal{L}_\text{dist} = d_H$ is the Hilbert projective distance. Therefore, we can establish a formal link between the Hilbert projective distance $d_H$ and the $\tilde{\mathcal{H}}$-divergence $d_{\tilde{\mathcal{H}}}$ provided in Theorem~\ref{thrm:old_upper_bound} by comparing both through the $TV$-divergence.
\begin{definition}[Total Variation Divergence] The total variation (TV) divergence, $d_\text{TV}$, between two distributions $\mathbb{P}$ and $\mathbb{P}'$ is defined as 
\begin{equation*}
    d_{\text{TV}}(\mathbb{P}, \mathbb{P}') := 2\sup_{B\in \mathcal{B}} | \mathbb{P}(B) - \mathbb{P}'(B)|
\end{equation*}
where $\mathcal{B}$ is the set of all measurable subsets under $\mathbb{P}$ and $\mathbb{P}'$.
\label{def:TV}
\end{definition}
In contrast to the common standard $d_{TV}$ distance~\citep{levin2017markov}, note that we keep the factor of 2 in Definition~\ref{def:TV} in analogy to \citep{cohen2024hyperboliccontractivityhilbertmetric}. 
\begin{remark}
From Definitions~\ref{def:H-div} and~\ref{def:TV}, it follows that $d_{\tilde{\mathcal{H}}} \leq d_{\text{TV}}$ because the supremum in the definition of $d_{\tilde{\mathcal{H}}}$ is taken only over the decision regions induced by $\tilde{\mathcal{H}}$, which is a subset of the collection of all measurable sets over which $d_{\text{TV}}$ takes its supremum.
\label{rmk:H-div_D-TV}
\end{remark}
\begin{proposition}\citep{cohen2024hyperboliccontractivityhilbertmetric} Given the probability distributions $\mathbb{P}$ and $\mathbb{P}'$, the $TV$ divergence is bounded by the Hilbert projective distance via the hyperbolic tangent function 
\begin{equation*}
        d_{\text{TV}}(\mathbb{P}, \mathbb{P}') \leq 2\tanh\frac{d_H(\mathbb{P}, \mathbb{P}')}{4}
\end{equation*}
\label{prop:d-TV_d-H}
\end{proposition}
\begin{proposition}[Upper Bound on Target Error] Given Remark~\ref{rmk:H-div_D-TV} and the established relation between $d_{\text{TV}}$ and $d_H$ in Proposition~\ref{prop:d-TV_d-H}, we can link $d_H$ and $d_{\tilde{\mathcal{H}}}$ for probability distributions $\mathbb{P}$ and $\mathbb{P}'$ as
\begin{equation*}
        d_{\tilde{\mathcal{H}}}(\mathbb{P}, \mathbb{P}') \leq 2\tanh\frac{d_H(\mathbb{P}, \mathbb{P}')}{4}
\end{equation*}
\label{prop:H-div_d-H}
\end{proposition}
Therefore, based on Proposition~\ref{prop:H-div_d-H}, we can rewrite the updated Theorem~\ref{thrm:old_upper_bound} with the Hibert projective distance.

\section{Experimental Details}
\label{app:exp_details}
\subsection{Image Denoising}
\label{app:exp_img_den}

\paragraph{Model.}
For the image denoising task, we adopt the exact autoencoder architecture described in \citet{9008549}. The encoder comprises three convolutional blocks followed by a linear layer of dimension $2$. Each block consists of a convolutional layer, a ReLU activation, and max pooling. The decoder mirrors this structure: a linear layer followed by three convolutional blocks, where max pooling is replaced with up-sampling operations to progressively reconstruct the input dimensionality. The full architecture is detailed in Table~16 of the Appendix in \citet{9008549}.

\paragraph{Data.}
We use MNIST and Fashion-MNIST, each originally split into $60{,}000$ train and $10{,}000$ test images. For both datasets, we partition each split evenly: half of the images are retained as clean source data, while the other half is corrupted to form the target domain. Following \citet{9008549}, we add Gaussian noise $N(0.4, 0.7^2)$ to all target images. This results in $30{,}000$ training samples per domain. From the source domain, we set aside $5{,}000$ images for validation, while evaluation is performed on $5{,}000$ unseen target-domain test samples. This protocol ensures no correspondence between source and target images.

\paragraph{Training.}
We closely follow the training configuration of \citet{9008549}. Specifically, we use a batch size of $128$, the Adam optimizer~\citep{kingma2014adam} with a fixed learning rate of $2\times 10^{-4}$, and train for $200$ epochs. The only tuned hyperparameter is $\beta$, which weights the adaptation loss. We select its value based on source-domain validation performance by searching over $\{0.1, 0.5, 1, 10, 10^2, \dots, 10^5\}$, and set $\beta=0.1$ in all reported experiments.

\subsection{Image Classification}
\label{app:exp_img_cls}

\paragraph{Model.}
Our backbone is ResNet-50 pretrained on ImageNet, a standard choice in prior UDA work. Following \citet{Chen_Fu_Chen_Jin_Cheng_Jin_Hua_2020}, we insert a bottleneck adaptation layer before the classifier. This adaptation layer is a fully connected layer of dimension $42$ for Office-31 and $25$ for VisDA-2017, followed by a $\tanh$ activation. Its output serves as input to the final classifier. The classifier itself is a linear layer of dimension $31$ for Office-31 and $12$ for VisDA-2017, matching the number of classes.

We set the hyperparameter $\eta=1$ for Office-31 without tuning. For VisDA-2017, monitoring the determinant of $P_S$ indicated that a smaller value was necessary to activate the adaptation mechanism, so we fixed $\eta=10^{-8}$.

\paragraph{Data.}
Office-31 contains three domains: Amazon ($2{,}817$ images), Webcam ($795$), and DSLR ($498$). VisDA-2017 contains three splits: train ($152{,}397$ images), validation ($55{,}388$), and test ($72{,}372$). Following \citet{Chen_Fu_Chen_Jin_Cheng_Jin_Hua_2020}, all images are resized to $224\times224$ pixels.

\paragraph{Training.}
To prevent rank-deficient covariance matrices, we balance batch size and feature dimensionality. A common heuristic requires at least ten times more samples than features. Accordingly, we use a batch size of $700$ for Office-31; for the DSLR domain (only $498$ images), we include all images in a single batch. For VisDA-2017, we set the batch size to $861$, the largest divisor of the train split size.

Consistent with \citet{Chen_Fu_Chen_Jin_Cheng_Jin_Hua_2020}, we fine-tune only the last convolutional layer for Office-31, and the last convolutional block for VisDA-2017, due to dataset size differences and limited computational resource available. In both cases, the adaptation and classifier layers are trained from scratch. We use the Adam optimizer with a learning rate of $3\times 10^{-5}$ for fine-tuned convolutional layers and $3\times 10^{-4}$ for newly initialized layers. Training runs for $1500$ epochs on Office-31 and up to $50$ epochs on VisDA-2017.

The adaptation weight $\beta$ is tuned per dataset. For Office-31, we select $\beta$ using the A$\rightarrow$W setup and use that value for all other setups, searching over $\{10^{-5}, 10^{-4}, \dots, 10^{-1}, 1\}$. For VisDA-2017, $\beta$ and the training epoch budget are chosen based on validation domain performance, searching over $\{10^{-2}, 10^{-1}, 1, 10\}$. The final settings are $\beta=10^{-3}$ for Office-31 and $\beta=10^{-1}$ for VisDA-2017.

\end{document}